\documentclass[manuscript,authorversion,nonacm]{acmart}

\AtBeginDocument{%
  \providecommand\BibTeX{{%
    \normalfont B\kern-0.5em{\scshape i\kern-0.25em b}\kern-0.8em\TeX}}}

\setcopyright{acmcopyright}
\copyrightyear{2023}
\acmYear{2023}
\acmDOI{XXXXXXX.XXXXXXX}

\usepackage{cleveref}

\acmConference[CSCW '23]{}{October 13,
  2023}{Minneapolis, Minnesota,, USA}
\acmBooktitle{Proceedings of the ACM on Human-Computer
Interaction, Vol. 6, CSCW1, (October 2023)}

\begin{document}


\title[Towards Meaningful Anomaly Detection]{Towards Meaningful Anomaly Detection: The Effect of Counterfactual Explanations on the Investigation of Anomalies in Multivariate Time Series}

\author{Max Schemmer}
\affiliation{%
  \institution{Karlsruhe Institute of Technology}
  \city{Karlsruhe}
  \country{Germany}
  }
\email{max.schemmer@kit.edu}

\author{Joshua Holstein}
\affiliation{%
  \institution{Karlsruhe Institute of Technology}
  \city{Karlsruhe}
  \state{Baden-Württemberg}
  \country{Germany}
  }
\email{joshua.holstein@.kit.edu}

\author{Niklas Bauer}
\affiliation{%
  \institution{Karlsruhe Institute of Technology}
  \city{Karlsruhe}
  \state{Baden-Württemberg}
  \country{Germany}
  }
\email{joshua.holstein@.kit.edu}

\author{Niklas Kühl}
\affiliation{%
  \institution{Karlsruhe Institute of Technology}
  \city{Karlsruhe}
  \country{Germany}
  }
\email{niklas.kuehl@kit.edu}

\author{Gerhard Satzger}
\affiliation{%
  \institution{Karlsruhe Institute of Technology}
  \city{Karlsruhe}
  \state{Baden-Württemberg}
  \country{Germany}
  }
\email{gerhard.satzger@.kit.edu}


\renewcommand{\shortauthors}{Schemmer et al.}

\begin{abstract}
Detecting rare events is essential in various fields, e.g., in cyber security or maintenance. Often, human experts are supported by anomaly detection systems as continuously monitoring the data is an error-prone and tedious task. 
However, among the anomalies detected may be events that are rare, e.g., a planned shutdown of a machine, but are not the actual event of interest, e.g., breakdowns of a machine.
Therefore, human experts are needed to validate whether the detected anomalies are relevant. We propose to support this anomaly investigation by providing explanations of anomaly detection. Related work only focuses on the technical implementation of explainable anomaly detection and neglects the subsequent human anomaly investigation.
To address this research gap, we conduct a behavioral experiment using records of taxi rides in New York City as a testbed. Participants are asked to differentiate extreme weather events from other anomalous events such as holidays or sporting events.
Our results show that providing counterfactual explanations do improve the investigation of anomalies, indicating potential for explainable anomaly detection in general.

\end{abstract}





\maketitle

\section{Introduction}
Detecting rare events is essential in various domains \citep{blazquez-garcia_review_2022,gamboa_deep_2017}. For example, in manufacturing, engineers want to find early indicators of machine failures that would allow them to retentively conduct maintenance. In cyber security, experts aim to find security breaches and attacks.
At the same time, monitoring continuous data streams is very challenging even for human experts \citep{qian_anomaly_2020,huang_cellular_2022}, primarily due to the vast amounts of data---in terms of granularity and variability---that need to be analyzed. 

For this reason, system designers build so-called anomaly detection systems (ADS) that aim to support human experts in identifying anomalies.
Recently, more and more of these systems are based on machine learning (ML) \citep{audibert_deep_2022, garg_evaluation_2022}, with autoencoder being the most commonly applied method \citep{chalapathy_deep_2019}.

\begin{figure}[h]
	\centering
	\includegraphics[width=0.5\textwidth]{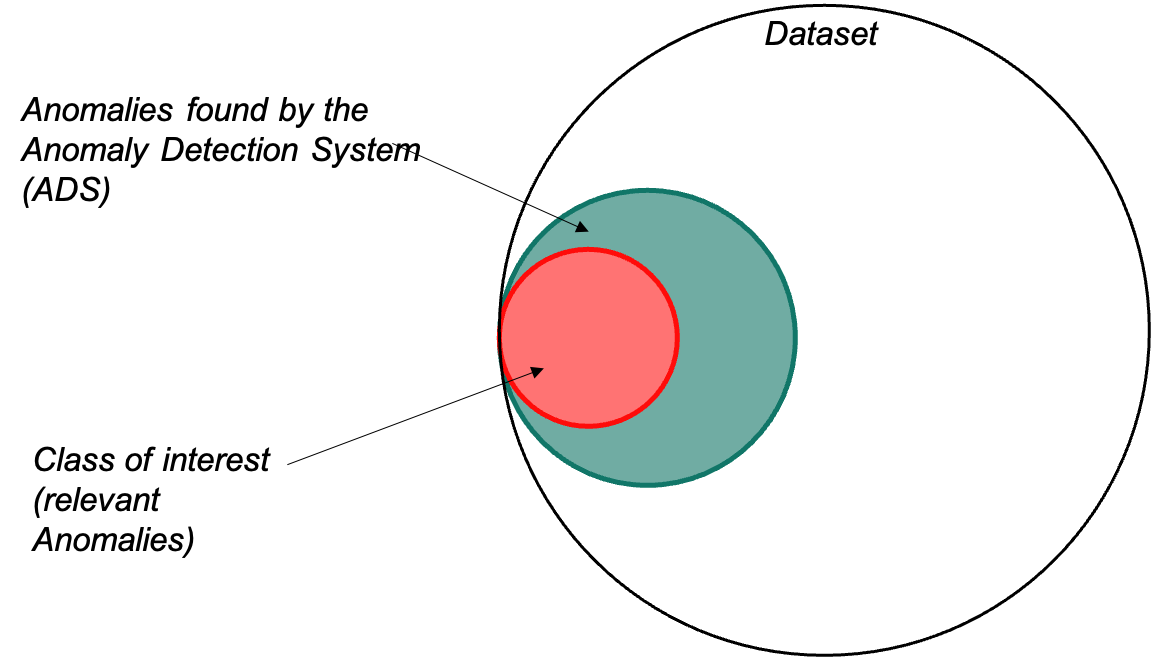}
	\caption{Abstract representation of rare event detection. The set of anomalies can be differentiated into relevant and non-relevant anomalies. In most use cases, domain experts are finally interested in detecting the class of interest.}
	\label{fig:incident-relvant-anomalies}
\end{figure}

However, anomaly detection in general and autoencoder, in particular,
cannot perform fully automated detection of \textit{relevant} anomalies. This is because ADS can only find anomalies, not the specific events that domain experts are interested in. Validating whether an identified anomaly is relevant is a challenging process that requires close collaboration between human experts and ADS. 
In manufacturing, for example, engineers are not interested in every anomaly in the production process but only those that might indicate machine failure. \Cref{fig:incident-relvant-anomalies} highlights this distinction between relevant and non-relevant anomalies.

To perform anomaly investigation, domain experts often need to investigate hundreds of different features \cite{liu_mtv_2022}, of which not all are relevant to a specific anomaly. For example, in maintenance, a typical production line registers thousands of measurements per second. To improve the accuracy of this classification process---the anomaly investigation---experts need support in boiling down the vast amount of interrelated data into relevant ones.

It becomes evident that the anomaly detection problem is actually twofold: detecting anomalies in the data and investigating them afterward.
While the first process step is well researched, the second lags behind \cite{chemweno2016rcam,steenwinckel2021flags,pang_deep_2022}. Even though it may not be possible to automatically investigate anomalies due to missing labels in many use cases, we hypothesize that ADS can still provide valuable information for human experts that improve their human classification. 
More precisely, we hypothesize that explanations of anomaly detection may support the subsequent classification task. In the manufacturing example, the ADS could highlight which sensor signals led to the detection of anomalies. This highlighting potentially simplifies the investigation of the anomaly by reducing the number of sensors that must be investigated.

A review of related literature reveals that while there are certainly studies focusing on explainable ADS \cite{CHOI2022109147, Fei2018}, none empirically evaluate whether these explanations support the subsequent anomaly investigation. Therefore, we derive the following research question:

\textbf{Research Question:} \textit{Can explanations of anomaly detection improve the anomaly investigation?}

To answer the research question, we conduct a behavioral experiment.
In this work, we focus on detecting and investigating anomalies in multivariate time series---a data type with many anomaly detection use cases \cite{Malhotra2016, kieu2019outlier}.
As the basis of our ADS, we have chosen an autoencoder as it is one of the most common approaches \cite{Malhotra2016}.
As previous research has pointed out, the most suitable type of explanation in the presence of multivariate time series might be counterfactual explanations \cite{ates_counterfactual_2021}. Counterfactual explanations are as similar as possible to the sample explained while having a different classification label; i.e., ``if the values of these particular time series were different in the given sample, the classification label would have been different.'' Other types of explanations exist, e.g., feature-based methods that highlight features relevant to the decision or example-based explanations that give an exemplary sample from the same class.
In the setting of time-series, \citet{ates_counterfactual_2021} argue that counterfactual explanations are superior to the previously mentioned explanation types as for feature importance, experts need to have knowledge of normal value ranges. Regarding example-based explanations, they argue that more than showing examples is required to understand the underlying patterns. We follow their thought process and focus on counterfactual explanations.
To summarize, we focus on counterfactual explanations for autoencoder in multivariate times series to investigate our research question.

Standard anomaly detection benchmark datasets for multivariate time series, e.g., from manufacturing \cite{bosch}, or cyber security \cite{Du2017}, have the drawback that they are too complex for behavioral online experiments. 
Therefore, based on a set of requirements, we searched for a suitable dataset and chose the New York City taxi trips. We use these recordings as a testbed and design an autoencoder that can identify different events by their substantial impact on the local taxi industry. As a subsequent classification task, we ask participants to differentiate extreme weather events from other events.
We conduct a behavioral experiment with 64 participants to answer our research question. We find that counterfactual explanations indeed improve the anomaly investigation.

To our knowledge, we are the first study that empirically investigates the effect of explanations of anomaly detection on anomaly investigation. By validating the potential, we motivate novel use cases based on anomaly detection that could have a major impact on how ADS are approached in the future.

Next, we will introduce the fundamentals of our work in \Cref{sec:related-work}. Following that, in  \Cref{sec:methodology}, we introduce, our dataset, the design of our explainable autoencoder and our experimental design. In \Cref{sec:results}, we present the results of our experiment. In \Cref{sec:discussion} we discuss our results, and in \Cref{sec:conclusion} we conclude.
\section{Related Work}
\label{sec:related-work}

In this chapter, we introduce the fundamentals of our work and provide an overview of related work.
First, we introduce foundations of anomalies, anomaly detection, investigation, and explainable AI. Then, we introduce the related work that covers explainable autoencoder-based anomaly detection in multivariate time series.

\subsection{Anomaly Definition.}
First, it is imperative to define the term ``anomaly'' to establish a common ground. An anomaly is essentially a data point or a sequence of data points with substantial deviations from the majority of data points \citep{hawkins_identification_1980,gornitz2013toward}. The term anomaly does not describe a specific event but rather a property of those events. Those events are described as ``unusual'' 
, ``rare'' and simply not ``normal'' \cite{gornitz2013toward}.

Anomalies can be categorized into different types:
Point anomalies are the most trivial to find, as these anomalies are only single points located outside the normal value range. Next, contextual anomalies can consist of sequences and can only be identified as anomalous in comparison to different points with the same context. The most complex type is the collective anomaly. Collective anomalies always span over sequences and only gradually show a different pattern compared to normal data. Individual values within this type of anomaly may seem ordinary and only collectively raise suspicion \cite{braei_anomaly_2020}. 

In most use cases where anomalies are to be detected, the ultimate goal is not to detect any anomaly, but particular ones \cite{song2007conditional,liu_mtv_2022}. 
Therefore, most anomaly detection use cases essentially boil down to a rare event classification. 
For example, in manufacturing, the goal of operators may be to detect early indicators of machinery failures. However, not only these early indicators but also other events deviate from the ``normal'' operation, e.g., planned shut-downs.
This means only a subset of the anomalies is actually of interest. 

To classify those rare events of interest, in the past knowledge-based systems were used, i.e. systems that explicitly store the knowledge of experts to detect the events \citep{steenwinckel2021flags}. However, experts have a limited, more global view, and as data size grows, it becomes harder for experts to explain deviations in values and their effects \citep{steenwinckel2021flags}. Moreover, acquiring their knowledge is a time-consuming and challenging task \citep{steenwinckel2021flags}.
For this reason, more and more ADS were developed. 
However, ADS cannot classify the detected anomalies based on their relevancy. This means a human anomaly investigation is still an imperative \cite{song2007conditional}.

In this work, we focus on improving this anomaly investigation with explanations of the anomaly detection.

\subsection{Anomaly Detection.}
Anomaly detection approaches consist of either classification (e.g., isolation forest), nearest neighbors (e.g., Distance Based), compression-based (e.g., autoencoder), or clustering methods \cite{muruti_survey_2018}. Further, anomaly detection can be categorized into three different classes \cite{Nassif2021}. \textit{Supervised anomaly detection} aims to build a classifier model that learns from a labeled training dataset. Here, the training dataset contains labels for normal and anomalous instances. In real-world, it may be challenging to create such datasets due to anomalies being rare events and models requiring vast amounts of data. Next, \textit{semi-supervised anomaly detection} requires only a training dataset with instances being labeled as normal. Accordingly, any instance different from the normal class is classified as an anomaly. Finally, \textit{unsupervised anomaly detection} is the most common type for anomaly detection as it does not require any labels in the training dataset, with autoencoders being one of the most powerful model classes.

While there are numerous methods to perform anomaly detection, the scope of this work is related to deep anomaly detection due to its superior performance \citep{chalapathy_deep_2019}. Deep anomaly detection describes the application of deep learning to the anomaly detection task. 
Autoencoders are the most frequently used deep anomaly detection method \citep{chalapathy_deep_2019}. This architecture was already introduced in the 1980's \citep{rumelhart_learning_1985} and attempts to compress the input data to then reconstruct it with as little information loss as possible \citep{baldi_autoencoders_2012}.
The most basic structure of an autoencoder contains an encoder that generates a compressed representation of the input data and a decoder that aims to reconstruct the input data from the compressed representation \citep{bank_autoencoders_2021}. Not all the information can be stored in the so-called bottleneck layer, so the model must learn statistical patterns in the training data \citep{bengio_learning_2009}. These lower-dimensional representations are the latent space of an autoencoder \citep{dillon_better_2021}. 
For time series anomaly detection, autoencoders are usually equipped with LSTM layers that can capture temporal dependencies \cite{Malhotra2016}. 
To decide whether a sample is anomalous or not the autoencoder reconstruction error is used. If the reconstruction error of a sample exceeds a certain threshold, the sample is labeled as an anomaly. The threshold can be tuned manually or set by a certain percentage of the highest errors.

\subsection{Anomaly Investigation.}
Only a few articles focus on anomaly investigation \citep{liu_mtv_2022,soldani2022anomaly}. 
Anomaly investigation can happen on multiple levels based on the necessary, and available data \cite{soldani2022anomaly}. It can either be conducted based on the same data used for the anomaly detection or by taking additional data into account. This work focuses on use cases where the same data is used.

Most of the existing work deals with data visualization to improve anomaly investigation \cite{soldani2022anomaly,xue_multivariate_2022}.
\citet{xue_multivariate_2022} develop an ADS for detecting and analyzing anomalies in cloud computing performance. They provide rich visualization and interaction designs to help understand the anomalies in a spatial and temporal context.
\citet{soldani2022anomaly} improve the process of detecting and investigating anomalies in time series data in industrial contexts. To do so, they characterized six design elements required and developed a visual ADS to support this process. 
They argue that future work should investigate how explanations can improve anomaly investigation. 

\subsection{Explainable Anomaly Detection}
Explanations are required to understand how specific predictions are generated \citep{simic_xai_2021}. 
On an abstract level, approaches can be divided into local and global explanations \citep{ates_counterfactual_2021}. Global explanations focus on the entire dataset \citep{ibrahim_global_2019}, whereas local explanations refer to individual observations \citep{plumb_model_2018}.

By reviewing related work, it becomes apparent that there are many implementations of explainable anomaly detection \cite{CHOI2022109147, Fei2018}. However, in most cases, the models were the only aspect evaluated quantitatively with metrics. 
Overall, to the best of our knowledge, no study has ever empirically evaluated whether the explanations of the anomaly detection also provide a benefit for anomaly investigation. 
Therefore, we argue that this work's topic is highly relevant.

\subsection{Explainable Autoencoder-based Anomaly Detection in Multivariate Time Series}

Within the context of multivariate time series, a lack of explainable AI approaches can be observed, while simultaneously, analytics for these time series are increasing in popularity \citep{ates_counterfactual_2021}.
Counterfactuals are a promising explainability technique for time series \citep{boubrahimi_mining_2022}.
While there are many counterfactual approaches in various domains, the multivariate time series domain remains mostly uncovered \citep{guidotti_counterfactual_2022}. 
Hereby, the work of \citet{ates_counterfactual_2021} is the only known framework for counterfactual explanations in time series classification. As their approach is model agnostic, they only require class probabilities as the model's output to create explanations. To do so, they modify the input data in a way that is as close as possible to the original input while receiving a different class label. However, not all available input features are altered and, instead, only the ones with the highest deviations between the original input and the modified instance with a different label. Reducing the number of adjusted variables helps human experts as previous research has pointed out that humans are only capable of processing four variables simultaneously \cite{Halford2005}. A typical example of counterfactual explanations outside the domain of time series is a loan application scenario: The AI declines a person's request, stating that similar customers have also been declined. In contrast, a counterfactual statement can convey that the request would have been accepted if the person had slightly lowered the credit amount \citep{kenny_generating_2020}.

During the review of related works that implement explainable autoencoder, it becomes apparent that most works utilize some form of feature importance as an explanation technique.
\cite{alfeo_using_2020, dix_three-step_2021, ghalehtaki_unsupervised_2022} use the model's built-in reconstruction error to detect important features. However, \citet{roelofs_autoencoder-based_2021} argues that this methodology is not very robust as the  reconstruction error does not always match the actual feature importance.
Other work uses well-known frameworks such as SHAP or LIME to generate feature importance through a surrogate model (e.g., \cite{jakubowski_explainable_2021}) or even deploy multiple SHAP explanations to capture temporal and feature interactions respectively \citep{hussain_explainable_2022}. \citet{ha_explainable_2022} calculates the feature importance through SHAP by applying a flattening layer on their LSTM autoencoder. The new model uses the weights from the autoencoder and generates explanations by using Gradient SHAP. \citet{oliveira_new_2022} designs its framework, the residual explainer, which interprets deviations of the reconstruction errors to create feature importance. In an experiment, the approach produces better results than SHAP and takes only a fraction of the time.
The only work to our knowledge that uses an explanation technique besides feature importance is the work of \citet{sulem_diverse_2022}, who generate counterfactual explanations.

To summarize, we do not find any study that empirically researched the influence of explainable anomaly detection on anomaly investigation.

\section{Methodology}
\label{sec:methodology}
In this section, we first derive hypotheses on the impact of explanations on anomaly investigation. Next, we provide information about the data and task we use to test the hypotheses. Then, we describe the development of the autoencoder and the counterfactual explanations, together forming our explainable ADS.
Finally, we present the experimental design.

\subsection{Hypotheses}
Anomaly investigation is often an error-prone and challenging task \cite{janus_applying_2021}. For example, in condition-based monitoring, experts are often left with an alarm and thousands of sensors which could lead to the anomaly being detected. 
Those explanations may have the potential to localize the anomalies, i.e., highlight the features relevant for the anomaly detection \cite{zimmerer2019unsupervised}. Further, in the case of counterfactual explanations, they may give context information on how typical values look.
Therefore, we hypothesize:

\textbf{H1:} {\itshape Providing explanations of the anomaly detection improves the effectiveness of the anomaly investigation.}

Additionally, explanations could not just improve the performance but also make the anomaly investigation faster and improve the efficiency of the process.
Therefore, we hypothesize:

\textbf{H2:} {\itshape Providing explanations of the anomaly detection improves the efficiency of the anomaly investigation.}

In addition, to formalizing hypothesizes on the direct effect of explanations on anomaly investigation, we discuss a first potential mediator of the effect---cognitive load \cite{plass2010cognitive}. Traditionally, cognitive load can be divided into the intrinsic cognitive load, i.e., difficulties of the task, and extraneous cognitive load, i.e., visualization of the task. In this work, we focus on extraneous cognitive load as intrinsic cognitive load cannot be changed \cite{plass2010cognitive}.
The explanations can help boil down from the potentially large amounts of variables to those relevant for anomaly investigation. Accordingly, fewer inter-variable effects must be considered, which leads to less extraneous cognitive load.
Therefore, we formulate:

\textbf{H3:} {\itshape Providing explanations of the anomaly detection decreases the extraneous cognitive load required for the anomaly investigation.}

Following related work \cite{Rao2020, GALY2012}, we hypothesize that cognitive load negatively influences anomaly investigation performance as well as efficiency:

\textbf{H4:} {\itshape Reduced extraneous cognitive load improves the effectiveness of the anomaly investigation.}

\textbf{H5:} {\itshape Reduced extraneous cognitive load improves the efficiency of the anomaly investigation.}

Having general hypotheses defined, in the following, we introduce our task and dataset to test the hypotheses.

\subsection{Dataset, Task \& Data Preprocessing  }

\textbf{Dataset.}
We search for a suitable task and dataset by specifying a list of requirements the dataset must fulfill.
The dataset must consist of multivariate time series and must include anomalies and, ideally, external information about the respective anomalies. Since the participants are non-experts, the dataset must come from a context they can understand. 

Based on these requirements, we evaluate several well-known multivariate benchmark datasets frequently used in anomaly detection on multivariate time series (e.g., \cite{bosch, Du2017}). All these datasets are multivariate and stem from a technical context. While these characteristics are desirable for a technical evaluation of a model, they conflict with our requirement to be easy to understand.

For this reason, we picked a dataset with a more common context.
One dataset that meets all these requirements is the public New York City Taxi dataset \cite{TLC}. 
Currently, around 1 million trips are recorded every day \cite{TLC}. TLC has made this data available to the public since 2009. Each trip record contains 19 features, e.g., information about the pick-up and drop-off time and location, the trip distance, payment types, fares, and the number of passengers.
Nearly 13 years of data are available - in these years, the taxi industry has changed considerably. Fares, availability of cabs, or, for example, new competitors have, among other factors, influenced the collected data and represent a considerable challenge that is out of the scope of this work. We address this issue by using a shorter period of observation. 

Certain days, such as holidays or days with extreme weather conditions, cause considerable deviations from the usual behavioral pattern. These days are thus suitable as anomalies because they are out-of-distribution by nature while serving as ground truth at the same time \citep{ferreira_visual_2013}. For extreme weather events, ground truth can be found on the governmental extreme weather website \footnote{\url{https://www.weather.gov/okx/stormevents}}. All of the anomalies are collective anomalies, e.g., they are just anomalous as a sequence.
During the chosen timeframe from the beginning of 2016 to the end of December 2018, several events with known large impacts on the taxi business took place, for example: 

\begin{itemize}
    \item Christmas (24.12 - 26.12.2018)
    \item New Year's Day (01.01.2018)
    \item Winter storm (15.11.2018
    \item Heavy snowfalls (21.03.2018)
\end{itemize}

\begin{figure}[h]
	\centering
	\includegraphics[width=1\textwidth]{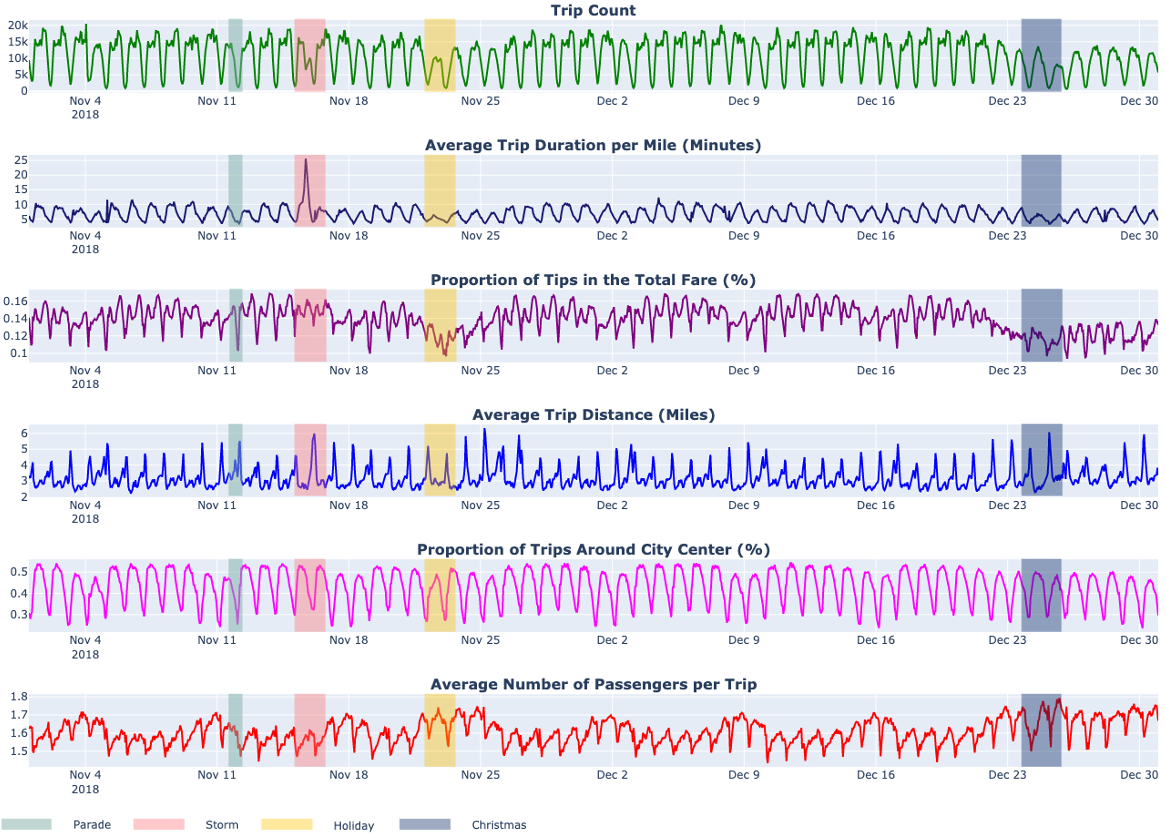}
	\caption{Excerpt of test data with highlighted event days.}
	\label{fig:test-data}
\end{figure}

For the training of our model, we use 2016 as train and 2017 as validation period. The year 2018 serves as our test set, of which we visualized a subset in Figure \ref{fig:test-data}. The colored areas shown indicate known events in New York City.

\textbf{Task.}
Accordingly, we want to also provide an easy-to-understand task based on the dataset that supports the classification of  the identified anomalies. Our dataset lays the ideal basis for this task as the anomalies have different classes, e.g., public holidays, extreme weather events, or other events. While it may not be possible to differentiate between such events deterministically, some frequently appearing patterns can be observed, e.g., during extreme weather, fewer people use taxis and, at the same time, the share of the tips increases while people usually give less tip on a holiday. Therefore, we provide participants with the task of classifying whether the shown anomaly is an extreme weather event or not. We employ a binary classification (extreme weather event or not) to be close to a realistic task. For example, in condition-based maintenance, the binary classification may be to discriminate faults from anomalies.

\textbf{Data Preprocessing.}
As the data is provided directly from the recording, it is necessary to clean and preprocess it. The goal of the preprocessing is to increase the data quality and, therefore, also the performance of the entire system \citep{frye_benchmarking_2021}. We merely make basic assumptions that ensure the validity of individual recordings while not removing any anomalies the model should detect, e.g., the trip duration should be longer than zero minutes \citep{institut_fur_wirtschaftsinformatik_und_digitale_gesellschaft_ev_potsdam_handling_2020}. 
After the data cleaning, we aggregate the taxi demand hourly and perform a few preprocessing steps. To increase the comprehensibility of the dataset, we drop some of the original 19 dimensions, as they are sometimes difficult to understand and negligible for anomaly detection. Further, we 
create additional features that are easy to interpret and thus support the classification task. Therefore, our final dataset consists of the following features: trip count, average trip duration per mile, the proportion of tips in the total fare, average trip distance, proportion of trips starting and ending in the city center, and, finally, the average number of passengers per trip.
Lastly, we scale the data to unify the magnitude of different features \citep{misra_impact_2019}, as this can have otherwise undesired results on the model's decision-making process \citep{pamir_non-technical_2022}.

\subsection{Explainable Anomaly Detection System}
\textbf{Modeling.} In the following, we present our anomaly detection modeling. 
Similar to related work \cite{ghalehtaki_unsupervised_2022, jakubowski_explainable_2021, ha_explainable_2022}, we use LSTM-layer to take intertemporal and multivariate dependencies into account. 
To optimize our architecture, we conduct a grid search and identify the following parameters as the best combination: window size (8), step size (2), hidden dimensions (8,6,4), and latent space (4). We use the reconstruction error and the detection of known anomalies as the target.


Our approach is to calculate the average reconstruction error of a window over all timesteps and features and compare this value to a threshold. The threshold must be optimized based on the results. 
The goal of this optimization is the recall of the model, meaning that all anomalies are identified as such by the model.

\textbf{Counterfactual Explanations.}
The standard autoencoder architecture must be extended to enable explanations for common explanation frameworks such as SHAP or CoMTE that cannot handle the autoencoder output. This is because the autoencoder output has the same dimensions as the input data. Current XAI frameworks, however, expect outputs in the form of a classification or regression prediction.
Thus, we design a new layer that manipulates the model's output to provide class probabilities 
\cite{ates_counterfactual_2021}.  
To calculate the necessary class probabilities, a new layer is given a threshold value in addition to the already existing sum of the reconstruction error, as proposed in \cite{ates_counterfactual_2021}. The threshold is determined by calculating the 99 \% percentile of the training data error. The reconstruction error is then, similar to \cite{ates_explainable_2020,aronsson_security_2021}, converted to a binary class probability by first subtracting the threshold value ($\tau$) from the calculated mean error (see Equation \ref{eq:proba}). The Sigmoid function afterward projects that value to a range between 0 and 1. The layer is concatenated after training the autoencoder.

\begin{equation}
((\frac{1}{N}\sum_{n=1}^{N}\lvert x_{n}-\hat{x}_{n}\rvert)-\tau)
\label{eq:proba}
\end{equation}

\begin{figure}[h]
	\centering
	\includegraphics[width=1\textwidth]{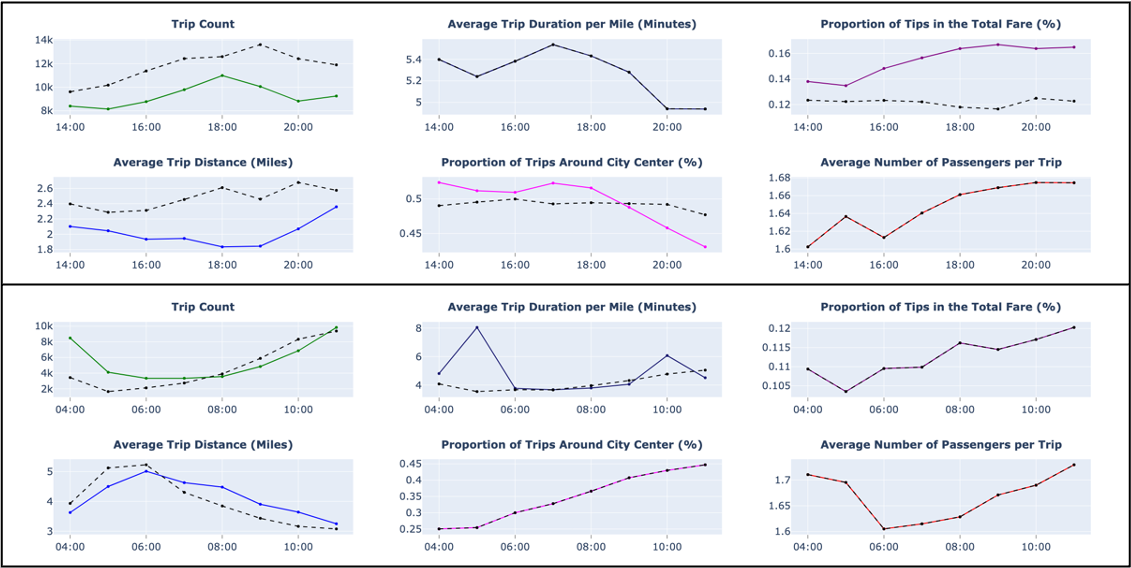}
	\caption{Exemplary explanations for extreme weather (top) and public holiday (bottom).}
	\label{fig:storm}
\end{figure}

Finally, we use the CoMTE framework to generate the counterfactual explanations \cite{ates_counterfactual_2021}, which serve two purposes.
First, they reduce the number of features that experts need to analyze (On average, our approach changes 3.2 features).
Second, they highlight how the time series should have looked liked to be not flagged as anomalous. \Cref{fig:storm} depicts two examples of our explanations. The used approach modifies four input features in the extreme weather event and three for the public holiday.

Having introduced our explainable ADS, we now describe how we conduct our behavioral experiment.

\subsection{Experimental Design}
\subsubsection{Pilot Study.}
To get qualitative feedback on our ADS, we conduct a focus group with five experts with backgrounds in machine learning.
The session lasted 30 minutes. First, we briefly introduce the basic information about this work and then focus on the core of the study. There, an anomalous window with the respective features is presented. 
The session is recorded and transcribed to evaluate the results more precisely in retrospect \citep{mcgrath_twelve_2019}.

The experts argue that the only cue generated by the explanations, the distance between the two lines of the counterfactual explanation, is not enough. We observe that it is vital to provide exemplary patterns in the pre-training of the user study to ensure that participants can understand the events being classified. We argue that this also transfers to real-world cases, as domain experts also have prior knowledge within their domain, which they incorporate into the anomaly investigation. 

\subsubsection{Study Procedure.}
The research model is tested in an online experiment with a between-subject design.
We tested two different conditions. First, a \textit{control} condition in which the human receives the ADS without counterfactual explanations, and second, a counterfactual explanation (\textit{CF}) condition. The study is approved by the University IRB.

\textbf{Sampling Strategy.}
In each condition, participants are provided with eight events. For the sampling of the eight events, we apply rules to ensure that the patterns of the underlying event are visible and prevent participants from being able to classify events based on previously seen anomalies, e.g., the same extreme weather at two different times during a day. 
Therefore, we first label our anomalies based on the provided dates, start and end times of extreme weather of the government storm website and public holidays \footnote{\url{https://publicholidays.com/us/new-york/2018-dates/}}. For extreme weather events, we label an anomaly as an extreme weather if the identified anomaly starts at most 2 hours before the start of the extreme weather or two hours before the end of the extreme weather. Similarly, we label anomalies as a holiday if they start on the date of a public holiday. Finally, we randomly draw four extreme weather events, three holiday events, and one anomaly classified as neither. While sampling, we verify that we do not draw two anomalies of the same day. 

\begin{figure}[h]
	\centering
	\includegraphics[width=0.9\textwidth]{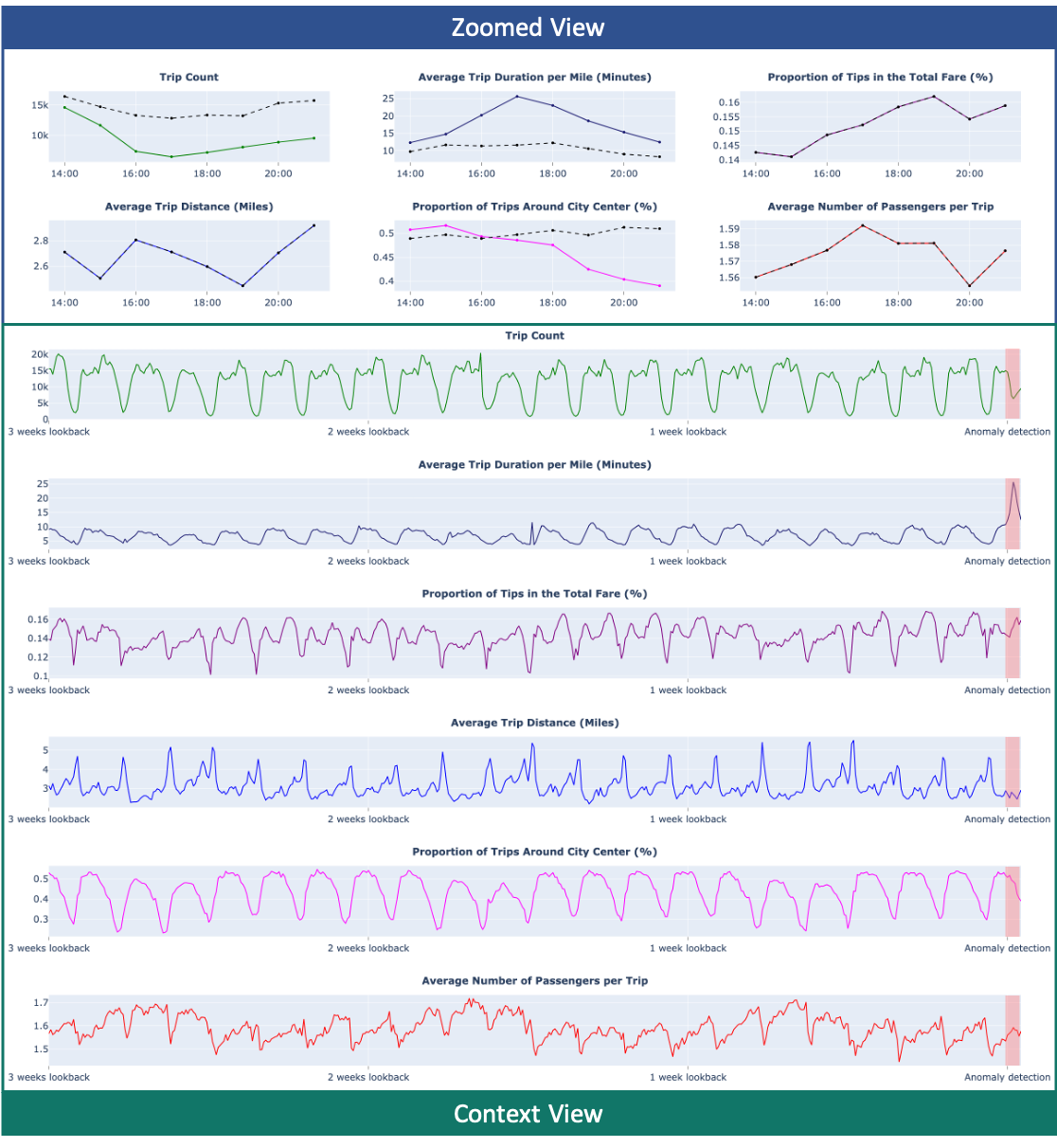}
	\caption{example of an extreme weather event with counterfactual explanations (top: zoomed view, bottom: context view).}
	\label{fig:interface}
\end{figure}

\textbf{Interface.} Next, we create visualizations of the sampled anomalies (see \Cref{fig:interface}). Similar to \citet{liu2021mtv}, we use two views with varying information: the context view and the zoomed view.
First, our context view shows past data of the last three weeks for all variables, with the anomaly being highlighted. This should support participants in understanding the behavior and interactions of the variables in non-anomalous times and thus provides context (following the requirements from the pilot study). 
However, we refrain from flagging additional anomalies in this period to avoid the possibility of inferring the date based on the position of the anomalies, e.g., New Year's Day, by the previous anomalies of Christmas.
Second, the zoomed view allows a detailed look at the specific time window of the anomaly. It is also the single point in which the treatments differ. While the AI treatment solely receives the data during the anomaly, the zoomed view of the counterfactual treatment additionally displays explanations.
To support the anomaly investigation, we provide supplemental information about the time of the anomaly to enable a better understanding of the anomaly's patterns without allowing conclusions on the specific date (e.g., Christmas).
For example, while an extreme weather event may result in fewer trips, this is also true for nights on regular working days.
Additionally, we argue that in real-world cases of anomaly, investigation time is a feature that is also available.

\textbf{Task flow.}
The online experiment is initiated with an attention control question that asks participants to state the color of grass. To control for internal validity, participants are randomly assigned to the condition groups.
As multivariate time series are difficult to interpret for humans \cite{janus_applying_2021}, we include multiple tutorials.
First, both conditions receive an introduction to the task and are given examples of extreme weather events and other events.
Following that, we explain the two views of our ADS and ask participants four comprehension questions.
Afterward, we give a short tutorial on how participants can detect extreme weather events, followed by two comprehension questions. Finally, we sample event patterns based on related literature \cite{qing2015identifying,lee2020large}.
For the CF condition, we follow on with an explanation of counterfactual explanations. We provide the participants with a general intuition of the explanations rather than specific technical information.
During the experiment, we neither used the terms AI and ML nor counterfactual explanations to prevent issues of AI literacy. 
Instead, we speak of ADS and expected values.
Then, the participants conduct two training tasks (one extreme weather event and one holiday) to familiarize the participants with the task and, depending on the condition, with its explanations. Additionally, the participants receive feedback on the training tasks. After the two training reviews, the participants are provided with the eight main tasks.  
For each task, we ask them how much they agree with the statement
``The anomaly is an extreme weather event'' on a four-point Likert scale (Strongly agree, agree, disagree, strongly disagree).
This allows us to get not only a binary classification but additionally certainty information.
After classifying the anomalies, we collect data on cognitive load and demographic variables.  

\textbf{Reward.}
To incentivize the participants, they were informed that for every correct decision, they get an additional 12 Cents in addition to a base payment of 6 Pounds per hour. However, the two training classifications do not count for the final evaluation. 

\textbf{Participant information.} 

The participants are recruited using the platform ``Prolific.co''.
We note that crowd workers might limit the generalizability of our results. However, our sampling of the task should ensure that crowd workers are capable of doing the task.
In total, we conducted the experiment with 66 participants (33 participants per condition).
We excluded two participants in the CF condition and two participants in the control condition because of conducting the eight tasks in under one minute. 
Apart from the attention check, we provided participants with in total of six questions that ensure that participants understand the task and underlying visualization, e.g., how many weeks of data are displayed in the context view. Based on these questions, we further excluded eight participants in the CF condition and nine participant in the control condition for incorrect answers. Even though this might seem like a high number of excluded participants, one needs to consider that multivariate time series anomaly detection is a very challenging task, and some crowd workers may not even understand what a time series is.
In addition, we exclude all participants who fall outside the interquartile range of 1.5. By doing so, we have excluded two outliers in the CF condition.
This leaves us with 22 participants in the control group and 21 in the counterfactual group.
\Cref{tab:Participants} shows the age, gender, and education distribution of the participants.

\begin{table}[H]
\caption{Summary of Participants' Characteristics.}
\label{tab:Participants}
\begin{tabular}{ll}
\toprule
Number per condition & Control = 22   \\
& Counterfactual Explanations = 21 \\
\midrule
Age & $\mu$ = 27.12, $\sigma$ = 6.29 \\
\midrule
Gender &  47 \% Female\\
 &  47 \% Male\\
  &  6 \% Non-Binary\\
 \midrule
 Education & 26 \% High school \\
 & 51 \% Bachelor  \\
 & 12 \% Master \\
 &  11 \% Other\\
\bottomrule
\end{tabular}
\end{table}

\subsubsection{Evaluation Measures.}

To evaluate our hypothesis, we calculate three measures based on the results of our experiment: effectiveness, efficiency, and, finally, extraneous cognitive load.

Our first measure, effectiveness, is the accuracy of the participants in the anomaly investigation, e.g., the share of correctly classified events. 
To calculate the share, we first binarize the result.
Due to our sampling strategy, by chance, participants would be able to have an accuracy of 50 percent.
In addition to the accuracy, we analyze the participants' certainty by comparing the share of agreement and disagreement with the percentage of strong agreement and disagreement.
Next, efficiency represents the time needed for the anomaly investigation. For both measures, we calculate the mean per participant for the global evaluation of our hypothesis. Additionally, we examine the effects of the explanation on each type of event more closely. Therefore, we also build the mean for the four extreme and non-extreme weather events.

Finally, we collect information about the extraneous cognitive load based on questions used by \citet{CHANG2017218}. It describes the cognitive load put on a person that is created via the presentation of the task. It is influenced, e.g., by the amount of irrelevant information displayed or unnecessarily presented content. Such information distracts from the relevant contents and thus leads to a higher extraneous cognitive load which hampers the task performance \cite{CHANG2017218}. To analyze our hypothesis, we build the mean per participant and compare it between treatments.

\section{Results}
\label{sec:results}
In the following section, we report the results of our study. First, we provide a qualitative interpretation of typical patterns of detected anomalies that could have been observed by the participants of the experiment alike. Finally, we present an analysis of the experiments' results. 

\subsection{Qualitative Interpretation of Detected Anomalies}

As mentioned earlier, experiment participants need to classify identified anomalies in extreme weather events. This classification is based on the intuition that each type of event has common patterns that are shared across anomalies. However, these patterns often do not allow a deterministic classification of anomalies. Nevertheless, in the following we qualitatively introduce and interpret certain patterns that were derived with the help of counterfactual explanations.

\textbf{Extreme Weather.} During extreme weather events, three variables mainly differ from regular days: trip count is lower, the proportion of tips in the total fare increases, and the average trip distance decreases. For an example of a winter storm, see the top of \Cref{fig:storm}. We interpret this pattern to mean that more people stay home on stormy days and forego longer trips, such as visiting relatives or friends in other city districts. Further, people leave their homes only for urgent matters and then rely on taxis. Once they have arrived at their destination, they express their gratitude to the taxi drivers with an increased tip because of the adverse circumstances.

\textbf{Public Holidays.} Compared to extreme weather events, public holidays often have a distinct pattern (see the bottom of \Cref{fig:storm}). On holidays the number of trips during the night is usually higher, and later, the average number of passengers per trip is higher
. People often go out the night before, and thus there are more trips during the night compared to regular days. Compared to regular working days, we interpret the higher trip count during the night with people that go out, which results in more trips. The second observation with more people sharing taxis could be families that visit relatives or friends together.

\subsection{Experiment Results}

In \Cref{tab:descriptives} we highlight the descriptive results of our experiment.
The results are split according to the experimental condition. Finally, we evaluate the significance of the results using the Student's T-tests or Mann-Whitney-U  tests after controlling for normality with the Shapiro–Wilk test. 
In the following, we first present our results regarding effectiveness and then highlight the impact of explanations on efficiency.

\begin{table}[H]
\caption{Descriptive outcomes.}
\label{tab:descriptives}
\begin{tabular}{lccccc}
\toprule
Condition  & Effectiveness & Efficiency  & Extraneous \\
& &  & Cognitive Load \\
\midrule
Control  & 55.11 \% (21.71 \%) & 30.92 s (22.2 s) & 3.21 (0.95)\\
Explanations  & 70.24 \% (15.04 \%) & 25.89 s (12.47 s) & 3.82 (1.41)\\
\bottomrule
\end{tabular}
\end{table}

\begin{figure}[]
	\centering
	\includegraphics[width=1\textwidth]{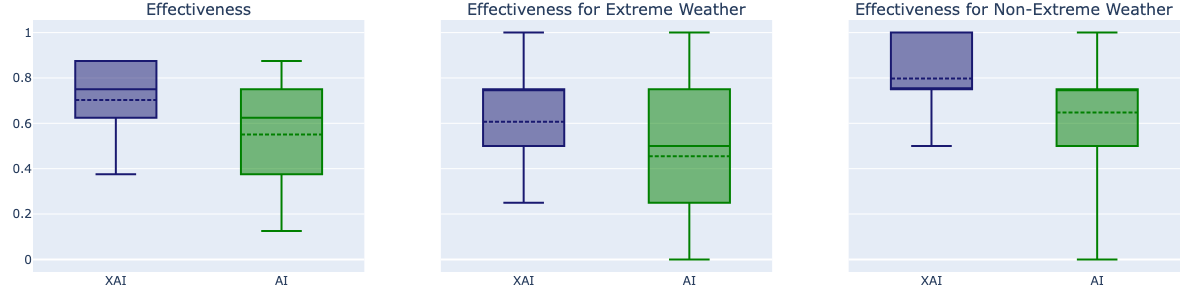}
	\caption{Distribution of the effectiveness of participants.}
	\label{fig:effectiveness}
\end{figure}

\textbf{Effectiveness.}
Analyzing the results of the experiment reveals that the participants' mean accuracy is significantly higher in the \textit{cf} group compared to the \textit{control} group  $(u =137.5 , p = 0.021)$. Thus, we can confirm H1 and conclude that \textbf{explanations improve the effectiveness of anomaly investigation}. A more detailed analysis based on the type of events reveals that the effectiveness increases for both: extreme weather events and non-extreme weather events. Individually applying Student's t-test to the effectiveness of each event type results in a p-value of 0.1 for only extreme weather events and also a p-value of 0.1 for non-extreme weather events. While these are not significant, it still provides a tendency that the increase in effectiveness draws from both event types alike, e.g., participants were not only better at detecting extreme weather events. Additionally,  \Cref{fig:effectiveness} shows that the interquartile range is lower for both events when provided with explanations.

\begin{figure}[]
	\centering
	\includegraphics[width=1\textwidth]{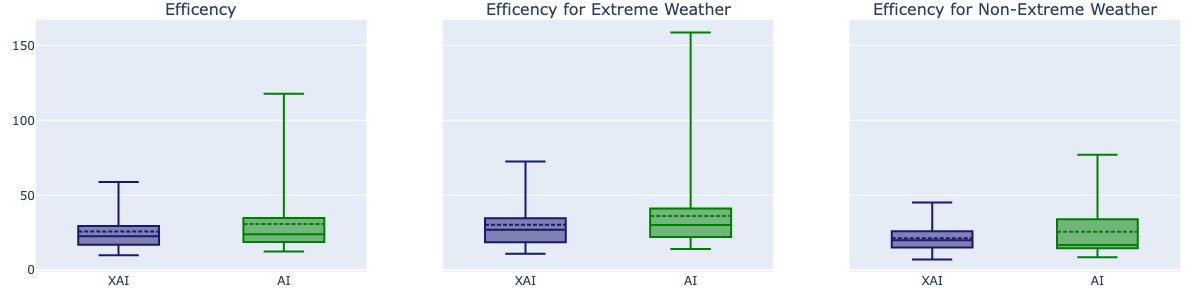}
	\caption{Distribution of the efficiency of participants.}
	\label{fig:efficency}
\end{figure}

\textbf{Efficiency.} Contrary to H2, counterfactual explanations do not increase human efficiency in anomaly investigation in our setup. Accordingly, no effect on the time needed for the classification was observable between the \textit{cf} group and the \textit{control} group. Neither further distinguishing between anomaly types (extreme weather vs. non-extreme weather)  nor task performance (correct vs. incorrect decision)  yielded significant effects on the efficiency of the participants. We conclude that we cannot verify H2. Next, we observed whether there are differences in efficiency between the type of events. As displayed in \Cref{fig:efficency}, the interquartile range is higher between treatments and for both kinds of anomalies considered individually.

\textbf{Cognitive Load.} Finally, we collect information on extraneous cognitive load to investigate a potential mediator of the effect. However, we do not find significant effects visible between treatments (H3). This means, we do not find evidence that explanations decrease cognitive load at anomaly investigation.

To investigate whether cognitive load in general, affects the anomaly investigation, we have a look at correlation effects between cognitive load and effectiveness/efficiency (H4/H5). As the Shapiro-Wilk test shows non-normality, we use the spearman correlation to determine the impact of cognitive load on participants' effectiveness and efficiency. The spearman test shows no significant correlation between task performance and cognitive load, which is why we can not confirm H4. However, further analysis shows a significant correlation between extraneous cognitive load and task efficiency, and, accordingly, we can confirm H5.

In total, our experiment highlights the potential for using counterfactual explanations to improve anomaly investigation. In the following section, we discuss our results.

\section{Discussion}
\label{sec:discussion}
In this work, we find that counterfactual explanations improve the accuracy of classifying weather events. This means human experts can transfer insights generated from an ADS to anomaly investigation (\textbf{H1}).

In our current setup, we do not find an efficiency improvement by providing counterfactual explanations (\textbf{H2}). This might be because counterfactual explanations also require effort to interpret.  

In addition to those direct effects, we investigate a first potential mediator---cognitive load. We find a trend that explanations do not reduce the cognitive load but instead increase it (\textbf{H3}). This might be induced due to counterfactual explanations being a nontrivial form of explanation that requires some cognitive effort to interpret. This means future research needs to investigate different potential mediators. Furthermore, future studies should identify the reasons for the increase in cognitive load and whether it can be mitigated, e.g., through a different form of visualization of counterfactual explanations.

\textbf{Implications.}
To the best of our knowledge, we are the first study empirically showing that explainable anomaly detection can improve anomaly investigation. This result has a major implication for research and practice.

Our work has implications for every use case with vast amounts of data and rare classes of interest. In previous work, these use cases were usually called disadvantageous for ML \cite{Guansong2021}. However, we argue they need a different solution approach. Instead of using supervised ML with advanced sampling strategies (e.g., SMOTE), we hypothesize that using explainable anomaly detection together with a human expert-based anomaly validation---i.e., human-AI collaboration--- can be superior.
.

In addition to detecting rare events of interest---thereby being an alarm system---explainable anomaly detection has the potential to work as a data mining tool to generate new knowledge. Anomaly detection can find patterns previously unknown to experts \cite{Chandola2009}. Explanations could enable experts to validate those patterns and thereby generate completely new insights.

\textbf{Limitations.}
As always with behavioral experiments, there is the question of how generalizable our results are. We would like to emphasize that our research does not aim to recommend generalizable design features (e.g., the use of counterfactual explanations for time series) but rather shows that explanation can improve the investigations of anomalies per se. 
We further argue that showing that it works with lay workers in an online experiment highlights even more potential for experts.
Still, additional work is needed to investigate design recommendations and whether our findings hold in other domains.

Furthermore, we want to discuss how realistic the testbed we have chosen is. For this reason, we compare the classification of taxi events with a typical use case for anomaly detection in manufacturing. 
Compared to our task, the number of features in the manufacturing industry is usually even higher. However, based on the researchers' expertise, we argue that the number of important features for the classification task is usually similarly small.

\textbf{Future Work.}
A key open question is how our human-AI collaboration workflow is perceived by end users. One key criterion for anomaly detection in the past was the reduction of ``false alarms'', i.e., detected anomalies that are not a class of interest, to reduce the work of experts \cite{Campos2016, Pang2019}. 
However, our results show that explanations may sometimes not reduce the time needed to interpret the anomalies. This means that explanations could be perceived as uncomfortable. Future research needs to investigate if experts perceive an explainable ADS as useful and will adopt it.

\section{Conclusion}
\label{sec:conclusion}
In this work, we address the problem of investigating anomalies regarding their relevancy.
In this research, we analyze the influence of explainable anomaly detection on anomaly investigation.
We conduct a behavioural experiment and show that counterfactual explanations of autoencoder-based anomaly detection improve investigating anomalies in multivariate time series.
We hope to motivate researchers and practitioners with our results to research, implement and use explainable anomaly detection.

\bibliographystyle{ACM-Reference-Format}
\bibliography{sample-base}

\end{document}